\documentclass[11pt,a4paper]{article}
\usepackage[hyperref]{acl2019}
\usepackage{times}
\usepackage{latexsym}

\usepackage{url}

\aclfinalcopy 

\usepackage{amsmath}
\usepackage{ifthen}
\usepackage{amssymb}
\usepackage{graphicx}
\usepackage{tabu}
\usepackage{booktabs}
\usepackage{subcaption}

\newcommand{\prob}[2][]{\text{\bf P}\ifthenelse{\not\equal{}{#1}}{_{#1}}{}\!\left(#2\right)}
\newcommand{\expect}[2][]{\text{\bf E}\ifthenelse{\not\equal{}{#1}}{_{#1}}{}\!\left[#2\right]}
\newcommand{\var}[2][]{\text{\bf Var}\ifthenelse{\not\equal{}{#1}}{_{#1}}{}\!\left[#2\right]}

\newcommand{\bx}{\mathbf{x}}

\newcommand{\by}{\mathbf{y}}

\newcommand{\ps}{p_{\mathrm{cm}}}
\newcommand{\pl}{p_{\mathrm{fm}}}
\newcommand{\cc}{C}
\newcommand{\qs}{q_{\mathrm{cm}}}

\newcommand{\softmax}{\mathrm{softmax}}

\newcommand{\ours}{Contextual Match}    

\title{Simple Unsupervised Summarization by Contextual Matching}

\author{Jiawei Zhou \\
  Harvard University \\ 
  \texttt{jzhou02@g.harvard.edu} \\\And
  Alexander M. Rush \\
  Harvard University \\ 
  \texttt{srush@seas.harvard.edu} \\}

\date{}

\begin{document}
\maketitle
\begin{abstract}
We propose an unsupervised method for sentence summarization using only language modeling.
The approach employs two language models, one that is generic (i.e. pretrained), and the other that is specific to the target domain.
We show that by using a product-of-experts criteria these are enough for  maintaining continuous contextual matching while 
maintaining output fluency. Experiments on both abstractive and extractive sentence summarization data sets show promising results of our method without being exposed to any paired data.

\end{abstract}

\section{Introduction}

Automatic text summarization is the process of formulating a shorter output text than the original while capturing its core meaning. 
We study the problem of unsupervised sentence summarization with no paired examples. While data-driven approaches have achieved great success based on various powerful learning frameworks such as sequence-to-sequence models with attention \cite{rush2015neural, chopra2016abstractive, nallapati2016abstractive}, variational auto-encoders \cite{miao2016language}, and reinforcement learning \cite{paulus2017deep}, they usually require a large amount of parallel data for supervision to do well. In comparison, the unsupervised approach reduces the human effort for collecting and annotating large amount of paired training data.

Recently researchers have begun to study the unsupervised sentence summarization tasks. 
These methods all use parameterized unsupervised learning methods to induce a latent variable model: for example 
\citet{schumann2018unsupervised} uses a length controlled variational autoencoder, \citet{fevry2018unsupervised} use a denoising autoencoder but only for extractive summarization, and \citet{wang2018learning} apply a reinforcement learning procedure combined with GANs, which takes a further step to the goal of  \citet{miao2016language} using language as latent representations for semi-supervised learning.

This work instead proposes a simple approach to this task that does not require any joint training. 
 We utilize a generic pretrained language model to enforce contextual matching between sentence prefixes. We then use a smoothed problem specific target language model to guide the fluency of the generation process. We combine these two models in a product-of-experts objective.
This approach does not require any task-specific training, yet experiments show results on par with or better than the best unsupervised systems while producing qualitatively fluent outputs. The key aspect of this technique is the use of a
pretrained language model for unsupervised contextual matching, i.e. unsupervised paraphrasing.

\section{Model Description}
Intuitively, a sentence summary is a shorter sentence that covers the main point succinctly. 
It should satisfy the following two properties (similar to \citet{pitler2010methods}): 
(a) Faithfulness: the sequence is close to the original sentence in terms of meaning;
(b) Fluency: the sequence is grammatical and sensible to the domain.

We propose to enforce the criteria using a product-of-experts model \cite{hinton2002training},
\begin{equation}\label{eq:modelbegin}
    \prob{\by|\bx} \propto \ps(\by|\bx)\pl(\by|\bx)^{\lambda},\quad |\by| \leq |\bx|
\end{equation}

\noindent where the left-hand side is the probability that a target sequence $\by$ is the summary of a source sequence $\bx$, $\ps(\by|\bx)$ measures the faithfulness in terms of contextual similarity from $\by$ to $\bx$, and $\pl(\by|\bx)$ measures the fluency of the token sequence $\by$ with respect to the target domain. We use $\lambda$ as a hyper-parameter to balance the two expert models.


We consider this distribution (\ref{eq:modelbegin}) being defined over all possible $\by$ whose tokens are restricted to a candidate list $\cc$ determined by $\bx$. For extractive summarization, $\cc$ is the set of word types in $\bx$. For abstractive summarization, $\cc$ consists of relevant word types to $\bx$ 
by taking $K$ closest word types from a full vocabulary $V$ for each source token measured by pretrained embeddings.

\subsection{Contextual Matching Model}

The first expert, $\ps(\by|\bx)$, tracks how  close $\by$ is to the original input $\bx$ in terms of a 
contextual ``trajectory''. We use a pretrained language model to define the left-contextual representations for both 
the source and target sequences. Define $S(x_{1:m}, y_{1:n})$ to be the contextual similarity between a source and target sequence of length $m$ and $n$ respectively under this model. We implement this as the cosine-similarity of a neural language model's final states with inputs $x_{1:m}$ and $y_{1:n}$. This approach relies heavily on the observed property that similar contextual sequences often correspond to paraphrases. If we can ensure close contextual matching, it will keep the output faithful to the original.

We use this similarity function to specify a generative process over the token sequence $\by$,
\[\ps(\by|\bx) = \prod_{n=1}^N \qs(y_n | \by_{<n}, \bx).\]
\noindent
The generative process aligns each 
target word to a source prefix. At the first step, $n=1$, we compute a greedy alignment score for each
possible word $w \in {\cal C}$, $s_w = \max_{j \geq 1} S(x_{1:j}, w)$ for all source prefixes up to length $j$. The probability $\qs(y_1 =w | \bx)$ is computed as $\softmax(\mathbf{s})$ over all target words. We also store the aligned context $z_1 = \arg\max_{j \geq 1} S(x_{1:j}, y_{1})$. 

For future words, we ensure that the alignment is strictly monotonic increasing, such that $z_n<z_{n+1}$ for all $n$. Monotonicity is a common assumption in summarization \cite{yu2016neural, yu2016online, raffel2017online}. For $n>1$ we compute the alignment score $s_w = \max_{j > z_{n-1}} S(x_{1:j}, [y_{1:n-1}, w])$ to only look at prefixes longer than $z_{n-1}$, the last greedy alignment. Since the distribution conditions on $\by_{}$ the past alignments are deterministic to compute (and can be stored). The main computational cost is in extending the target language model context to compute $S$. 


\begin{figure}
    \centering
    \includegraphics[width=0.45\textwidth]{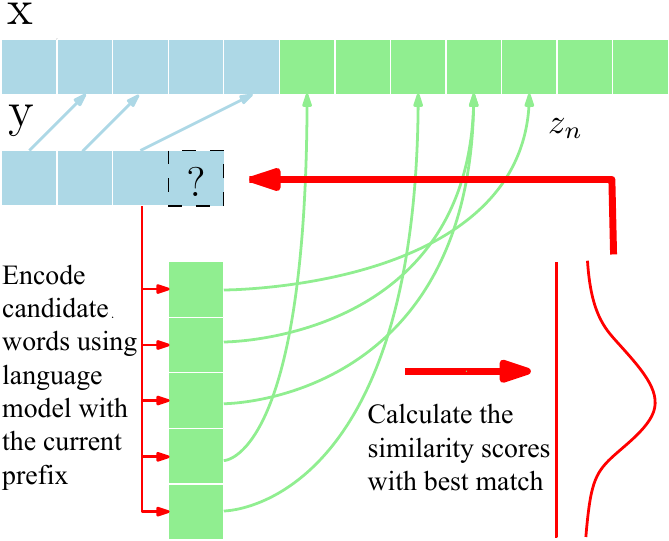}
    \caption{Generative process of the contextual matching model.}
    \label{fig:ps}
\end{figure}

This process is terminated when a sampled token in $\by$ is aligned to the end of the source sequence $\bx$, and the strict monotonic increasing alignment constraint guarantees that the target sequence will not be longer than the source sequence. The generative process of the above model is illustrated in Fig.~\ref{fig:ps}.

\subsection{Domain Fluency Model}

The second expert, $\pl(\by|\bx)$, accounts for the  fluency of $\by$ with respect to the target domain. It directly is based on a domain specific language model. Its role is to adapt the output to read closer shorter sentences common to the summarization domain. Note that unlike the contextual matching model where $\by$ explicitly depends on $\bx$ in its generative process, in the domain fluency language model, the dependency of $\by$ on $\bx$ is implicit through the candidate set $C$ that is determined by the specific source sequence $\bx$.

The main technical challenge is that the probabilities of a pretrained language model are not well-calibrated with the contextual matching model within the candidate set $C$, and so the language model tends to dominate the objective because it has much lower variance (more peaky) in the output distribution than the contextual matching model. To manage this issue we apply kernel smoothing over the language model to adapt it from the full vocab $V$ down to the candidate word list $C$. 

Our smoothing process focuses on the output embeddings from the pretrained language model. First we form the Voronoi partition \cite{aurenhammer1991voronoi} over all the embeddings using the candidate set $C$. That is, each word type $w'$ in the full vocabulary $V$ is exactly assigned to one region represented by a word type $w$ in the candidate set $C$, such that the distance from $w'$ to $w$ is not greater than its distance to any other word types in $C$. As above, we use cosine similarity between corresponding word embeddings to define the regions. This results in a partition of the full vocabulary space into $|C|$ distinct regions, called Voronoi cells. For each word type $w\in C$, we define ${\cal N}(w)$ to be the Voronoi cell formed around it.
 We then use cluster smoothing to define a new probability distribution:
\begin{equation*}
    \pl(\by | \bx)=\prod_{n=1}^N \sum_{w'\in\mathcal{N}(y_n)}\mathrm{lm}(w'|\by_{<n})
\end{equation*}
\noindent
where $\mathrm{lm}$ is the conditional probability distribution of the pretrained domain fluency language model.
By our construction, $\pl$ is a valid distribution over the candidate list $C$.
The main benefit is that it redistributes probability mass lost to terms in $V$ to the active words in $C$. 
We find this approach smoothing balances integration with $\ps$. 

\subsection{Summary Generation}

To generate summaries we maximize the log probability (\ref{eq:modelbegin}) to approximate $\by^*$ using beam search. We begin with a special start token.
A sequence is moved out of beam if it has aligned to the end token appended to the source sequence. To discourage extremely short sequences, we apply length normalization to re-rank the finished hypotheses. We choose a simple length penalty as $lp(\by)=|\by| + \alpha$ with $\alpha$ a tuning parameter.

\section{Experimental Setup}
For the contextual matching model's similarity function $S$, we adopt the forward language model of ELMo \cite{peters2018deep} to encode tokens to corresponding hidden states in the sequence, resulting in a three-layer representation each of dimension 512. The bottom layer is a fixed character embedding layer, and the above two layers are LSTMs associated with the generic unsupervised language model trained on a large amount of text data.
We explicitly manage the ELMo hidden states to allow our model to generate contextual embeddings sequentially for efficient beam search.\footnote{Code available at https://github.com/jzhou316/Unsuper
vised-Sentence-Summarization.}
The fluency language model component $\mathrm{lm}$ is task specific, and pretrained on a corpus of summarizations. We use an LSTM model with 2 layers, both embedding size and hidden size set to 1024. It is trained using dropout rate 0.5 and SGD combined with gradient clipping.

We test our method on both abstractive and extractive sentence summarization tasks.
For abstractive summarization, we use the English Gigaword data set pre-processed by \citet{rush2015neural}. 
We train $\pl$ using its 3.8 million headlines in the training set, and generate summaries for the input in test set.
For extractive summarization, we use the Google data set from \citet{filippova2013overcoming}. We train $\pl$ on 200K compressed sentences in the training set and test on the first 1000 pairs of evaluation set consistent with previous works.
For generation, we set $\lambda=0.11$ in (\ref{eq:modelbegin}) 
and beam size to 10.
Each source sentence is tokenized and lowercased, with periods deleted and a special end of sentence token appended. In abstractive summarization, we use $K=6$ in the candidate list and use the fixed embeddings at the bottom layer of ELMo language model for similarity.
Larger $K$ has only small impact on performance but makes the generation more expensive. The hyper-parameter $\alpha$ for length penalty ranges from -0.1 to 0.1 for different tasks, mainly for desired output length as we find ROUGE scores are not sensitive to it.
We use concatenation of all ELMo layers as default in $\ps$.


\section{Results and Analysis}

\textbf{Quantitative Results.} 
The automatic evaluation scores are presented in Table \ref{table:abstractive} and Table \ref{table:extractive}. 
For abstractive sentence summarization,
we report the ROUGE F1 scores compared with baselines and previous unsupervised methods.
Our method outperforms commonly used prefix baselines for this task which take the first 75 characters or 8 words of the source as a summary. 
Our system achieves comparable results to \citet{wang2018learning} a system based on both GANs and reinforcement training.
Note that the GAN-based system needs both source and target sentences for training (they are unpaired), whereas our method only needs the target domain sentences for a simple language model.
In Table~\ref{table:abstractive}, we also list scores of the state-of-the-art supervised model, an attention based seq-to-seq model of our own implementation,
as well as the oracle scores of our method obtained by choosing the best summary among all finished hypothesis from beam search. The oracle scores are much higher, indicating that our unsupervised method does allow summaries of better quality, but with no supervision it is hard to pick them out with any unsupervised metric.
For extractive sentence summarization, our method achieves good compression rate and significantly raises a previous unsupervised baseline on token level F1 score. 

\begin{table}[t!]

\begin{center}
\begin{tabular}{lccc}  
\toprule
Model & R1 & R2 & RL \\ 
\midrule
Lead-75C & 23.69  & 7.93 & 21.5 \\
Lead-8 & 21.30 & 7.34 & 19.94 \\
\midrule
\citet{schumann2018unsupervised} & 22.19 & 4.56 & 19.88\\
\citet{wang2018learning} & 27.09 & 9.86 & 24.97 \\
\ours\ & 26.48 & 10.05 & 24.41 \\
\midrule
\citet{cao2018retrieve} & 37.04 & 19.03 & 34.46 \\
seq2seq & 33.50 & 15.85 & 31.44 \\
\midrule
Contextual Oracle & 37.03 & 15.46 & 33.23 \\
\bottomrule
\end{tabular}
\end{center}
\caption{Experimental results of abstractive summarization on Gigaword test set with ROUGE metric. The top section is prefix baselines, the second section is recent unsupervised methods and ours, the third section is state-of-the-art supervised method along with our implementation of a seq-to-seq model with attention, and the bottom section is our model's oracle performance. \citet{wang2018learning} is by author correspondence (scores differ because of evaluation setup). For another unsupervised work \citet{fevry2018unsupervised}, we attempted to replicate on our 
test set, but were unable to obtain results better than the baselines.}
\label{table:abstractive}
\end{table}

\begin{table}[t!]

\begin{center}
\begin{tabular}{lcc}
\toprule
Model & F1 & CR \\ 
\midrule
F\&A Unsupervised   & 52.3  &  - \\
\ours & 60.90  & 0.38\\
\midrule
\citet{filippova2015sentence} & 82.0 & 0.38\\
\citet{zhao2018language} & 85.1 & 0.39 \\
\midrule
Contextual Oracle & 82.1 & 0.39 \\
\bottomrule
\end{tabular}
\end{center}
\caption{Experimental results of extractive summarization on Google data set. F1 is the token overlapping score, and CR is the compression rate. F\&A is an unsupervised baseline used in \citet{filippova2013overcoming}, and the middle section is supervised results.}
\label{table:extractive}
\end{table}

\begin{table}[t!]
\small
\begin{center}
\begin{tabular}{l ccc cc}
\toprule
 & \multicolumn{3}{c}{abstractive} & \multicolumn{2}{c}{extractive}\\
Models & R1 & R2 & RL & F1 & CR \\ 
\midrule
CS + cat & 26.48 & 10.05 & 24.41 & 60.90 & 0.38\\
CS + avg & 26.34 & 9.79 & 24.23 & 60.09 & 0.38\\
CS + top & 26.21 & 9.69 & 24.14 & 62.18 & 0.34\\
CS + mid & 25.46 & 9.39 & 23.34 & 59.32 & 0.40\\
CS + bot & 15.29 & 3.95 & 14.06 & 21.14 & 0.23\\
\midrule
TEMP5 + cat & 26.31 & 9.38 & 23.60 & 52.10 & 0.43\\
TEMP10 + cat & 25.63 & 8.82 & 22.86 & 42.33 & 0.47\\
NA + cat & 24.81 & 8.89  & 22.87 & 49.80 & 0.32\\
\bottomrule
\end{tabular}
\end{center}
\caption{Comparison of different model choices. The top section evaluates the effects of contextual representation in the matching model, and the bottom section evaluates the effects of different smoothing methods in the fluency model.}
\label{table:analysis}
\end{table}

\paragraph{Analysis.}
Table \ref{table:analysis} considers analysis of different aspects of the model. First, we look at the 
fluency model and compare the cluster smoothing (CS) approach with softmax temperature (TEMPx with x being the temperature) commonly used for generation in LM-integrated models \cite{chorowski2016towards} as well as no adjustment (NA). Second, we vary the 3-layer representation out of ELMo forward language model to do contextual matching (bot/mid/top: bottom/middle/top layer only, avg: average of 3 layers, cat: concatenation of all layers). 

Results show the effectiveness of our cluster smoothing method for the vocabulary adaptive language model $\pl$, although temperature smoothing is an option for abstractive datasets. Additionally Contextual embeddings have a huge impact on performance. 
 When using word embeddings (bottom layer only from ELMo language model) in our contextual matching model $\ps$, the summarization performance drops significantly to below simple baselines as demonstrated by score decrease. This is strong evidence that encoding independent tokens in a sequence with generic language model hidden states helps maintain the contextual flow. 
 Experiments also show that even when only using $\ps$ (by setting $\lambda=0$), utilizing the ELMo language model states allows the generated sequence to follow the source $\bx$ closely, whereas normal context-free word embeddings would fail to do so.

\begin{table}[t!]
\small
\begin{center}
\begin{tabular}{p{0.45\textwidth}}
\toprule
I: japan 's nec corp. and UNK computer corp. of the united states said wednesday they had agreed to join forces in supercomputer sales\\
G: nec UNK in computer sales tie-up \\
s2s: nec UNK to join forces in supercomputer sales \\
GAN: nec corp. to join forces in sales \\
CM (cat): nec agrees to join forces in supercomputer sales\\
CM\ (top): nec agrees to join forces in computer sales\\
CM\ (bot): nec to join forces in supercomputer sales\\
\midrule
I: turnout was heavy for parliamentary elections monday in trinidad and tobago after a month of intensive campaigning throughout the country , one of the most prosperous in the caribbean\\
G: trinidad and tobago poll draws heavy turnout by john babb\\
s2s: turnout heavy for parliamentary elections in trinidad and tobago\\
GAN: heavy turnout for parliamentary elections in trinidad\\
CM (cat): parliamentary elections monday in trinidad and tobago\\
CM\ (top): turnout is hefty for parliamentary elections in trinidad and tobago\\
CM\ (bot): trinidad and tobago most prosperous in the caribbean\\
\midrule
I: a consortium led by us investment bank goldman sachs thursday increased its takeover offer of associated british ports holdings , the biggest port operator in britain , after being threatened with a possible rival bid\\
G: goldman sachs increases bid for ab ports\\
s2s: goldman sachs ups takeover offer of british ports\\
GAN: us investment bank increased takeover offer of british ports\\
CM (cat): us investment bank goldman sachs increases shareholdings\\
CM\ (top): investment bank goldman sachs increases investment in britain\\
CM\ (bot): britain being threatened with a possible bid\\
\bottomrule
\end{tabular}
\end{center}
\caption{Abstractive sentence summary examples on Gigaword test set. I is the input, G is the reference, s2s is a supervised attention based seq-to-seq model, GAN is the unsupervised system from \citet{wang2018learning}, and CM is our unsupervised model. The third example is a failure case we picked where the sentence is fluent and makes sense but misses the point as a summary.}
\label{table:examples}
\end{table}

\begin{figure}[t!]
    \centering
    \begin{subfigure}[t]{0.5\textwidth}
    \includegraphics[width=\textwidth]{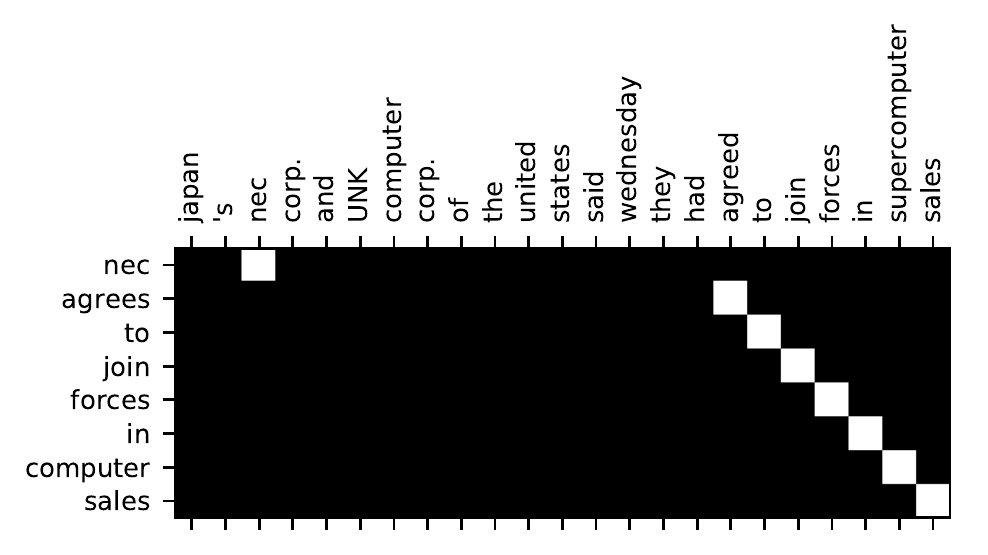}
    \end{subfigure}
    \begin{subfigure}[t]{0.5\textwidth}
    \includegraphics[width=\textwidth]{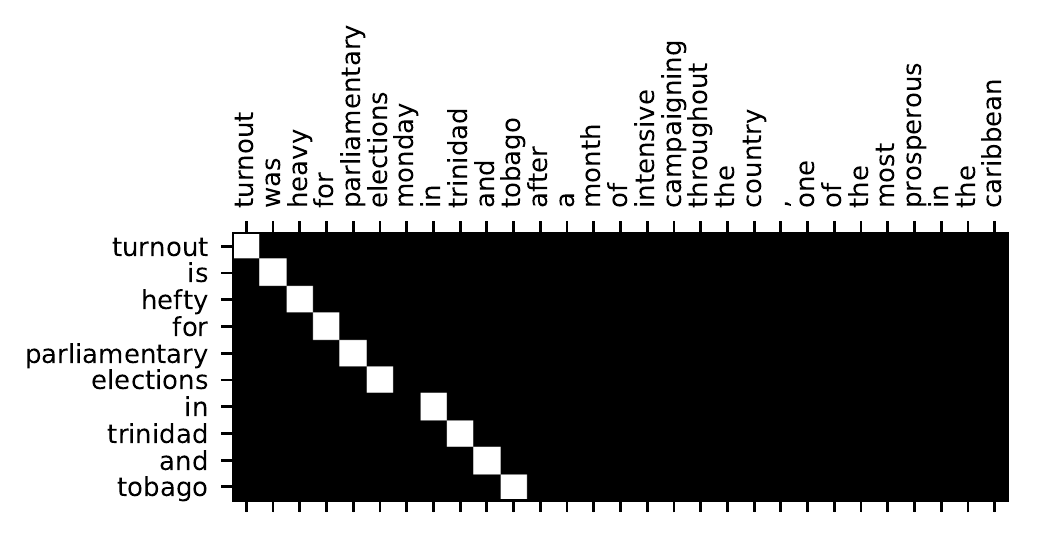}
    \end{subfigure}
    \begin{subfigure}[t]{0.5\textwidth}
    \includegraphics[width=\textwidth]{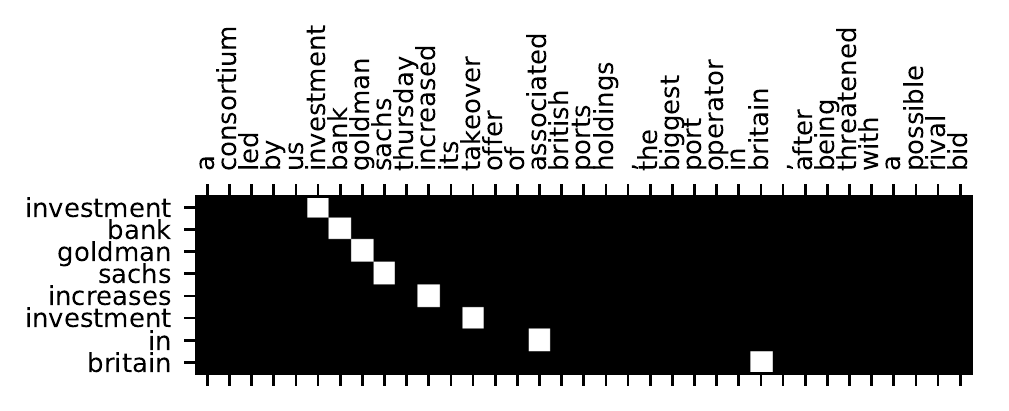}
    \end{subfigure}
    \caption{Examples of alignment results generated by our unsupervised method between the abstractive summaries and corresponding source sentences in the Gigaword test set.}
    \label{fig:alignment}
\end{figure}

Table \ref{table:examples} shows some examples of our unsupervised generation of summaries, compared with the human reference, an attention based seq-to-seq model we trained using all the Gigaword parallel data, and the GAN-based unsupervised system from \citet{wang2018learning}.
Besides our default of using all ELMo layers, we also show generations by using the top and bottom (context-independent) layer only.
Our generation has fairly good qualities, and it can correct verb tenses and paraphrase automatically. Note that top representation actually finds more abstractive summaries (such as in example 2),
and the bottom representation fails to focus on the proper context.
The failed examples are mostly due to missing the main point,
as in example 3, 
or the summary needs to reorder tokens in the source sequence. 
Moreover, as a byproduct, our unsupervised method naturally generates hard alignments between summary and source sentences in the contextual matching process. We show some examples in Figure~\ref{fig:alignment} corresponding to the sentences in Table~\ref{table:examples}.

\section{Conclusion}

We propose a novel methodology for unsupervised sentence summarization using contextual matching. Previous neural unsupervised works mostly adopt complex encoder-decoder frameworks. 
We achieve good generation qualities and competitive evaluation scores. We also demonstrate a new way of utilizing pre-trained generic language models for contextual matching in untrained generation. Future work could be comparing language models of different types and scales in this direction.

\section*{Acknowledgements}

We would like to thank Yuntian Deng and Yoon Kim
for useful discussions. This work was supported 
by NSF 1845664 and research awards from Google, Oracle, and Facebook.


\bibliography{acl2019}
\bibliographystyle{acl_natbib}

\end{document}